\documentclass[a4paper]{llncs}

\usepackage{svg}
\usepackage{csquotes}
\usepackage{hyperref}

\usepackage{subfig}

\usepackage{amssymb}
\usepackage{colortbl}
\usepackage{multirow}
\usepackage{footnote}
\usepackage{algorithm}
\usepackage{algpseudocode} 
\usepackage{cite}
\usepackage{bbding}

\makesavenoteenv{tabular}
\makesavenoteenv{table}

\begin{document}

	\title{The Dreaming Variational Autoencoder for Reinforcement Learning Environments}
	\titlerunning{The Dreaming Variational Autoencoder}

	\author{Per-Arne Andersen\textsuperscript{(\Envelope)} \and Morten Goodwin \and Ole-Christoffer Granmo}
	\authorrunning{P. Andersen et al.}
	\institute{Department of ICT, University of Agder, Grimstad, Norway\\
		\email{\{per.andersen,morten.goodwin,ole.granmo\}@uia.no}}
	\maketitle              
	\begin{abstract}
		Reinforcement learning has shown great potential in generalizing over raw sensory data using only a single neural network for value optimization. There are several challenges in the current state-of-the-art reinforcement learning algorithms that prevent them from converging towards the global optima. It is likely that the solution to these problems lies in short- and long-term planning, exploration and memory management for reinforcement learning algorithms. Games are often used to benchmark reinforcement learning algorithms as they provide a flexible, reproducible, and easy to control environment. Regardless, few games feature a state-space where results in exploration, memory, and planning are easily perceived. This paper presents \textit{The Dreaming Variational Autoencoder} (DVAE), a neural network based generative modeling architecture for exploration in environments with sparse feedback. We further present Deep Maze, a novel and flexible maze engine that challenges DVAE in partial and fully-observable state-spaces, long-horizon tasks, and deterministic and stochastic problems. We show initial findings and encourage further work in reinforcement learning driven by generative exploration.
		
		\keywords{Deep Reinforcement Learning  \and Environment Modeling \and Neural Networks \and Variational Autoencoder \and Markov Decision Processes \and Exploration \and Artificial Experience-Replay}
		
	\end{abstract}

	\section{Introduction}
	Reinforcement learning (RL) is a field of research that has quickly become one of the most promising branches of machine learning algorithms to solve artificial general intelligence~\cite{Kaelbling1996, Li2017, Arulkumaran2017, Mousavi2018}. There have been several breakthroughs in reinforcement learning in recent years for relatively simple environments \cite{Mnih2013, Mnih2015, Chen2015, VanSeijen2017}, but no algorithms are capable of human performance in situations where complex policies must be learned. Due to this, a number of open research questions remain in reinforcement learning. It is possible that many of the problems can be resolved with algorithms that adequately accounts for planning, exploration, and memory at different time-horizons.

	In current state-of-the-art RL algorithms, long-horizon RL tasks are difficult to master because there is as of yet no optimal exploration algorithm that is capable of proper state-space pruning. Exploration strategies such as $\epsilon$-greedy is widely used in RL, but cannot find an adequate exploration/exploitation balance without significant hyperparameter-tuning. Environment modeling is a promising exploration technique where the goal is for the model to imitate the behavior of the target environment. This limits the required interaction with the target environment, enabling nearly unlimited access to exploration without the cost of exhausting the target environment. In addition to environment-modeling, a balance between exploration and exploitation must be accounted for, and it is, therefore, essential for the environment model to receive feedback from the RL agent.
	
	By combining the ideas of variational autoencoders with deep RL agents, we find that it is possible for agents to learn optimal policies using only generated training data samples. The approach is presented as the dreaming variational autoencoder. We also show a new learning environment, Deep Maze, that aims to bring a vast set of challenges for reinforcement learning algorithms and is the environment used for testing the DVAE algorithm.
	
	This paper is organized as follows. 
	Section \ref{sec:bg} briefly introduces the reader to preliminaries. Section \ref{sec:dvae} proposes \textit{The Dreaming Variational Autoencoder} for environment modeling to improve exploration in RL. Section \ref{sec:environments} introduces the Deep Maze learning environment for exploration, planning and memory management research for reinforcement learning.  Section \ref{sec:experiments} shows results in the Deep Line Wars environment and that RL agents can be trained to navigate through the deep maze environment using only artificial training data.

	
	
	
	
	
	
	
	\section{Related Work}
	In machine learning, the goal is to create an algorithm that is capable of constructing a model of some environment accurately. There is, however, little research in \textit{game} environment modeling in the scale we propose in this paper. The primary focus of recent RL research has been on the value and policy aspect of RL algorithm, while less attention has been put into perfecting environment modeling methods.
	
	In 2016, the work in \cite{Bangaru2016} proposed a method of deducing the Markov Decision Process (MDP) by introducing an adaptive exploration signal (pseudo-reward), which was obtained using deep generative model. Their method was to compute the Jacobian of each state and used it as the pseudo-reward when using deep neural networks to learn the state-generalization.
	
    Xiao et al. proposed in \cite{Xiao2016} the use of generative adversarial networks (GAN) for model-based reinforcement learning. The goal was to utilize GAN for learning dynamics of the environment in a short-horizon timespan and combine this with the strength of far-horizon value iteration RL algorithms. The GAN architecture proposed illustrated near authentic generated images giving comparable results to \cite{Mnih2013}.  
    
    In \cite{Higgins2017} Higgins et al. proposed DARLA, an architecture for modeling the environment using $\beta$-VAE \cite{Higgins2016}. The trained model was used to extract the optimal policy of the environment using algorithms such as DQN~\cite{Mnih2015}, A3C~\cite{Mnih2016}, and Episodic Control~\cite{Blundell2016}. DARLA is to the best of our knowledge, the first algorithm to properly introduce learning without access to the target environment during training.
    
  	Buesing et al. recently compared several methods of environment modeling, showing that it is far better to model the state-space then to utilize Monte-Carlo rollouts (RAR). The proposed architecture, state-space models (SSM) was significantly faster and produced acceptable results compared to auto-regressive (AR) methods. \cite{Buesing2018}
  	
  	Ha and Schmidhuber  proposed in \cite{Ha2018} \textit{World Models}, a novel architecture for training RL algorithms using variational autoencoders. This paper showed that agents could successfully learn the environment dynamics and use this as an exploration technique requiring no interaction with the target domain.
    
	\section{Background}
	\label{sec:bg}
	We base our work on the well-established theory of reinforcement learning and formulate the problem as a MDP \cite{Sutton1998}. 
	An MDP contains $(\mathcal{S}, \mathcal{A}, \mathcal{T}, r)$ pairs that define the environment as a model. The state-space, $\mathcal{S}$ represents all possible states while the action-space, $\mathcal{A}$ represents all available actions the agent can perform in the environment. $\mathcal{T}$ denotes the transition function $(\mathcal{T}: \mathcal{S} \times \mathcal{A} \rightarrow \mathcal{S})$, which is a mapping from state $s_{t} \in \mathcal{S}$ and action $a_{t} \in \mathcal{A}$ to the future state $s_{t+1}$. After each performed action, the environment dispatches a reward signal, $\mathcal{R}: \mathcal{S} \rightarrow r$.
	
	We call a sequence of states and actions a \textit{trajectory} denoted as \\\(\tau = (s_{0}, a_{0},\dots,s_{t},a_{t})\) and the sequence is sampled through the use of a stochastic policy that predicts the optimal action in any state: $\pi_{\theta}(a_{t}|s_{t})$, where $\pi$ is the policy and $\theta$ are the parameters. The primary goal of the reinforcement learning is to \textit{reinforce} good behavior. The algorithm should try to learn the policy that maximizes the total expected discounted reward given by, $\mathcal{J}(\pi) = \mathop{\mathbb{E}}_{(s_{t}, a_{t}){\sim}p(\pi)}\big[\sum_{i=0}^{T}\gamma^{i}\mathcal{R }(s_{i})\big]$ \cite{Mnih2015}.
	
	\section{The Dreaming Variational Autoencoder}
	\label{sec:dvae}
	The Dreaming Variational Autoencoder (DVAE) is an end-to-end solution for generating probable future states \( \hat{s}_{t+n}\) from an arbitrary state-space \( \mathcal{S} \) using state-action pairs explored prior to \(s_{t+n}\) and \(a_{t+n}\).

	\begin{figure}
	\centering
	\includegraphics[width=0.5\linewidth]{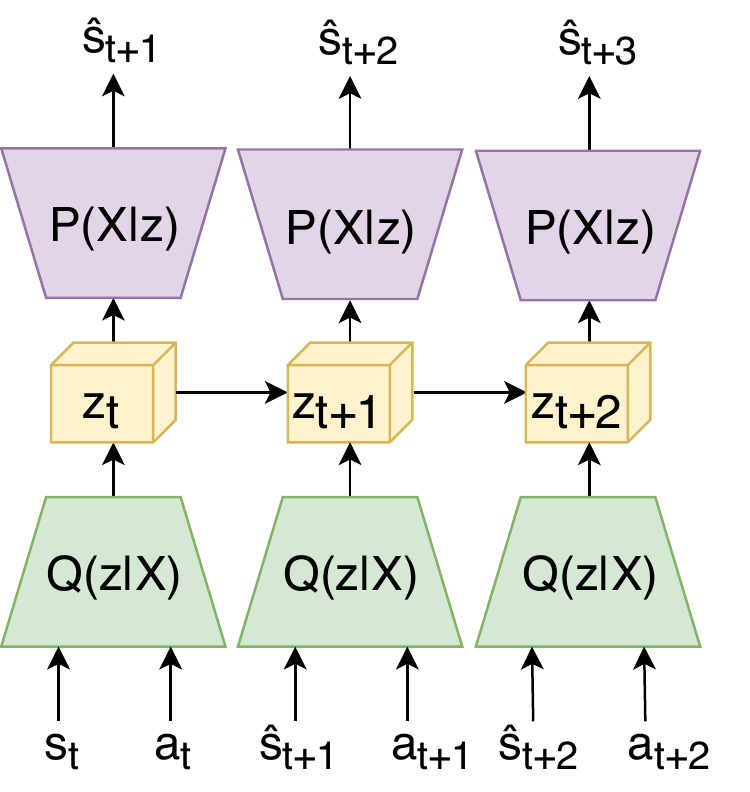}
	\caption{Illustration of the DVAE model. The model consumes state and action pairs, yielding the input encoded in latent-space. Latent-space can then be decoded to a probable future state. $\mathcal{Q}(z|X)$ is the encoder, $z_{t}$ is latent-space, and $\mathcal{P}(X|z)$ is the decoder. DVAE can also use LSTM to better learn longer sequences in continuous state-spaces.}
	\label{fig:dvae_model}
\end{figure}

	\begin{algorithm}[!htp]
		\caption{The Dreaming Variational Autoencoder} 
		\label{alg1} 
		\begin{algorithmic}[1]
			\State Initialize replay memory \( \mathcal{D} \) and \( \mathcal{\hat{D}}\) to capacity \( \mathcal{N} \)
			\State Initialize policy $\pi_{\theta}$ 
			
			\Function{Run-Agent}{$\mathcal{T}$, $\mathcal{D}$}
			
			\For {i = 0 to N\_EPISODES}
			\State Observe starting state, $s_{0} \sim \mathcal{N}(0,1)$
			
			\While{$s_{t}$ not TERMINAL}
			
			\State $a_{t} \gets \pi_{\theta}(s_{t} = s)$
			\State $s_{t+1}, r_{t}, terminal_{t} \gets \mathcal{T}(s_{t}, a_{t})$	
			\State Store experience into replay buffer $\mathcal{D}(s_{t}, a_{t}, r_{t}, s_{t+1}, terminal_{t})$
			\State $s_{t} \gets s_{t+1}$
			\EndWhile
			\EndFor
			
			\EndFunction
			
			\State Initialize encoder $\mathcal{Q}(z|X)$
			\State Initialize decoder $\mathcal{P}(X|z)$
			\State Initialize DVAE model $\mathcal{T}_{\theta} = \mathcal{P}(X|\mathcal{Q}(z|X))$
			
			\Function{DVAE}{}
			
			\For{$d_{i}$ in D}
			
			\State $s_{t}, a_{t}, r_{t}, s_{t+1} \gets d_{i}$\Comment{Expand replay buffer pair}
			
			\State $X_{t} \gets s_{t}, a_{t}$
			\State $z_{t} \gets \mathcal{Q}(X_{t})$ \Comment{Encode $X_{t}$ into latent-space}
			\State $\hat{s}_{t+1} \gets \mathcal{P}(z_{t})$\Comment{Decode $z_{t}$ into probable future state}
			\State Store experience into artificial replay buffer $\mathcal{\hat{D}}$($\hat{s}_{t}, a_{t}, r_{t}, \hat{s}_{t+1}, terminal_{t}$)
			\State $\hat{s}_{t} = \hat{s}_{t+1}$				
			\EndFor
			
			\State \textbf{return} $\mathcal{\hat{D}}$
			
			\EndFunction
		\end{algorithmic}
	\end{algorithm}

	The DVAE algorithm, seen in Figure~\ref{fig:dvae_model} works as follows.
	First, the agent collects experiences for utilizing experience-replay in the \textit{Run-Agent}  function. At this stage, the agent explores the state-space guided by a Gaussian distributed policy. The agent acts, observes, and stores the observations into the experience-replay buffer $\mathcal{D}$. After the agent reaches terminal state, the DVAE algorithm encodes state-action pairs from the replay-buffer $\mathit{D}$ into probable future states. This is stored in the replay-buffer for artificial future-states $\hat{D}$.

	\begin{table}[!htp]
		\centering
		\caption{DVAE algorithm for generating states using $\mathcal{T}_{\theta}$ versus the real transition function $\mathcal{T}$. First, a real state is collected from the replay-memory. DVAE can then produce new states from current the trajectory $\tau$ using the state-action pairs. $\theta$ represent the trainable model parameters.}
		\begin{tabular}{lllllllll}
			&                                                  &                                                  &                                         & \multicolumn{2}{l}{}                                                                                &                                           & \multicolumn{2}{l}{}                                                                                \\
			& \cellcolor[HTML]{F4C2C2}{\color[HTML]{F4C2C2} 1} & \cellcolor[HTML]{333333}{\color[HTML]{333333} 0} &                                         & \cellcolor[HTML]{333333}{\color[HTML]{333333} 0} & \cellcolor[HTML]{F4C2C2}{\color[HTML]{F4C2C2} 1} &                                           & \cellcolor[HTML]{333333}{\color[HTML]{333333} 0} & \cellcolor[HTML]{333333}{\color[HTML]{333333} 0} \\
			\multirow{-2}{*}{Real States}                          & \cellcolor[HTML]{333333}{\color[HTML]{333333} 0} & \cellcolor[HTML]{333333}{\color[HTML]{333333} 0} & \multirow{-2}{*}{$\mathcal{T}(s_{0}, \mathcal{A}_{right})$} & \cellcolor[HTML]{333333}{\color[HTML]{333333} 0} & \cellcolor[HTML]{333333}{\color[HTML]{333333} 0} & \multirow{-2}{*}{$\mathcal{T}(s_{1}, \mathcal{A}_{down})$}    & \cellcolor[HTML]{333333}{\color[HTML]{333333} 0} & \cellcolor[HTML]{F4C2C2}{\color[HTML]{F4C2C2} 1} \\
			& $s_{0}$                                          &                                                  &                                         & \multicolumn{2}{l}{$s_{1}$}                                                                         &                                           & \multicolumn{2}{l}{$s_{2}$}                                                                         \\ \hline
			&                                                  &                                                  &                                         & \multicolumn{2}{l}{}                                                                                &                                           & \multicolumn{2}{l}{}                                                                                \\ \cline{2-3}
			\multicolumn{1}{l|}{}                                  & \multicolumn{2}{c|}{}                                                                               &                                         & \cellcolor[HTML]{333333}{\color[HTML]{333333} 0} & \cellcolor[HTML]{F4C2C2}{\color[HTML]{F4C2C2} 1} &                                           & \cellcolor[HTML]{333333}{\color[HTML]{333333} 0} & \cellcolor[HTML]{333333}{\color[HTML]{333333} 0} \\
			\multicolumn{1}{l|}{\multirow{-2}{*}{Generated States}} & \multicolumn{2}{c|}{\multirow{-2}{*}{N/A}}                                                          & \multirow{-2}{*}{$\mathcal{T}_{\theta}(s_{0},\mathcal{A}_{right}, \theta)$}  & \cellcolor[HTML]{333333}{\color[HTML]{333333} 0} & \cellcolor[HTML]{333333}{\color[HTML]{333333} 0} & \multirow{-2}{*}{$\mathcal{T}_{\theta}(\hat{s}_1,\mathcal{A}_{down}, \theta)$} & \cellcolor[HTML]{333333}{\color[HTML]{333333} 0} & \cellcolor[HTML]{F4C2C2}{\color[HTML]{F4C2C2} 1} \\ \cline{2-3}
			&                                                  &                                                  &                                         & \multicolumn{2}{l}{$\hat{s}_1$}                                                                     &                                           & \multicolumn{2}{l}{$\hat{s}_2$}                                                                    
		\end{tabular}
		\label{tbl:dvae_state_gen}
	\end{table}
		
	Table \ref{tbl:dvae_state_gen} illustrates how the algorithm can generate sequences of artificial trajectories using $\mathcal{T}_{\theta} = \mathcal{P}(X|\mathcal{Q}(z|X))$, where $z = \mathcal{Q}(z|X)$ is the encoder, and $\mathcal{T}_{\theta} = \mathcal{P}(X|z)$ is the decoder. With state $s_0$ and action $\mathcal{A}_{right}$ as input, the algorithm generates state $\hat{s}_1$ which in the table can be observed is similar to the real state $s_1$. With the next input, $\mathcal{A}_{down}$, the DVAE algorithm generates the next state $\hat{s}_2$ which again can be observed to be equal to $s_2$. Note that this is without ever observing state $s_1$. Hence, the DVAE algorithm needs to be initiated with a state, e.g. $s_0$, and actions follows. It then generates (dreams) next states, 
	
	The requirement is that the environment must be partially discovered so that the algorithm can learn to behave similarly to the target environment. To predict a trajectory of three timesteps, the algorithm does nesting to generate the whole sequence: $\tau = \hat{s}_{1},a_{1},\hat{s}_{2},a_{2}, \hat{s}_{3},a_{3} = \mathcal{T}_{\theta}(\mathcal{T}_{\theta}(\mathcal{T}_{\theta}(s_{0}, \mathcal{A}_{rnd}), \mathcal{A}_{rnd}), \mathcal{A}_{rnd})$. The algorithm does this well in early on, but have difficulties with long sequences beyond eight in continuous environments.
	
	\section{Environments}
	\label{sec:environments}
	The DVAE algorithm was tested on two game environments. The first environment is Deep Line Wars \cite{Andersen2017}, a simplified Real-Time Strategy game. We introduce Deep Maze, a flexible environment with a wide range of challenges suited for reinforcement learning research.
	
	\subsection{The Deep Maze Environment}
	\label{sec:deep_maze}
	
	   \begin{figure}
		\subfloat[A Small, Fully Observable MDP]{%
			\includegraphics[width=0.49\linewidth]{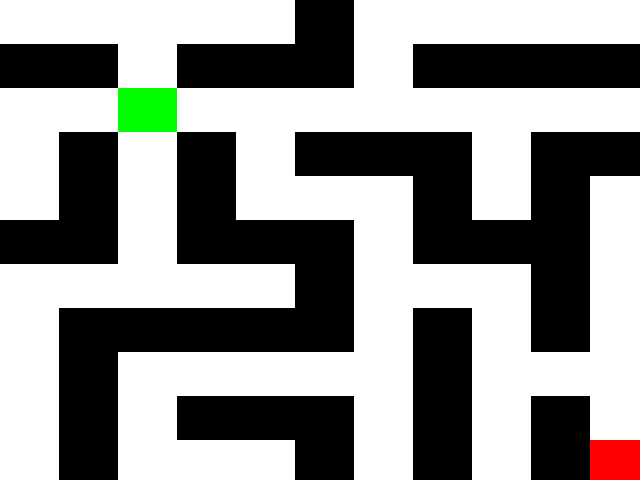}
		}
		\hfill
		\subfloat[A Large, Fully Observable MDP]{%
			\includegraphics[width=0.49\linewidth]{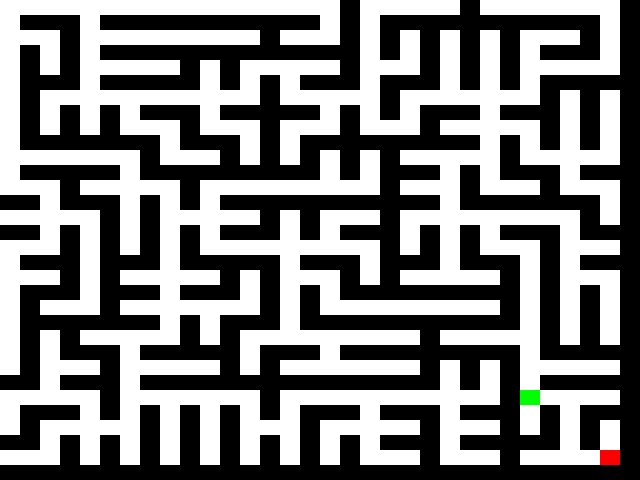}
		}
		\hfill
		\vfill
	
		\subfloat[Partially Observable MDP having a vision distance of 3 tiles]{%
			\includegraphics[width=0.49\linewidth]{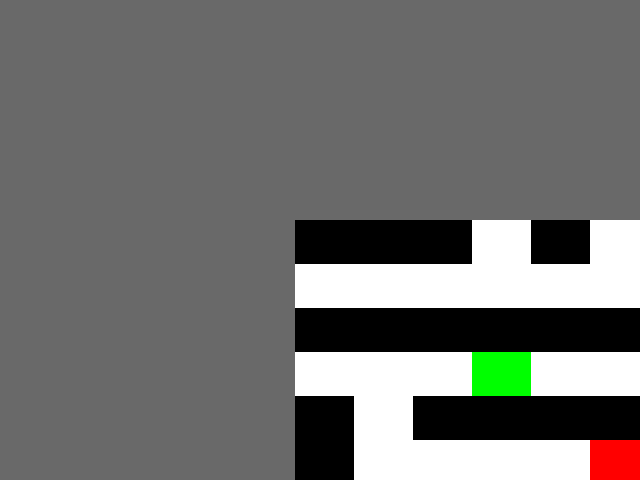}
		}
		\hfill
		\subfloat[Partially Observable MDP having ray-traced vision]{%
			\includegraphics[width=0.49\linewidth]{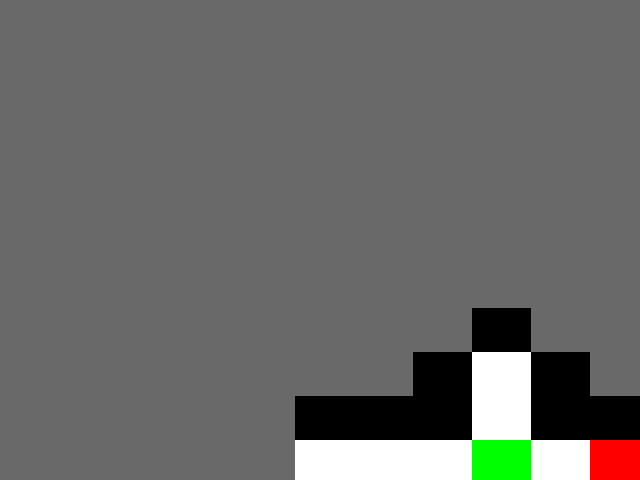}
		}
		
		\caption{Overview of four distinct MDP scenarios using Deep Maze.}
		\label{fig:deep_maze}
	\end{figure}


	
	The Deep Maze is a flexible learning environment for controlled research in exploration, planning, and memory for reinforcement learning algorithms. Maze solving is a well-known problem, and is used heavily throughout the RL literature \cite{Sutton1998}, but is often limited to small and fully-observable scenarios. The Deep Maze environment extends the maze problem to over 540 unique scenarios including Partially-Observable Markov Decision Processes (POMDP). Figure \ref{fig:deep_maze} illustrates a small subset of the available environments for Deep Maze, ranging from small-scale MDP's to large-scale POMDP's. The Deep Maze further features custom game mechanics such as relocated exits and dynamically changing mazes.
	
	The game engine is modularized and has an API that enables a flexible tool set for third-party scenarios. This extends the capabilities of Deep Maze to support nearly all possible scenario combination in the realm of maze solving.\footnote{The Deep Maze is open-source and publicly available at \url{https://github.com/CAIR/deep-maze}.}
	
	\subsubsection{State Representation}
	RL agents depend on sensory input to evaluate and predict the best action at current timestep. Preprocessing of data is essential so that agents can extract features from the input. For this reason, Deep Maze has built-in state representation for RGB Images, Grayscale Images, and raw state matrices.  
	
	\subsubsection{Scenario Setup}
    The Deep Maze learning environment ships with four scenario modes: (1) Normal, (2) POMDP, (3) Limited POMDP, and (4) Timed Limited POMDP. 

	The first mode exposes a seed-based randomly generated maze where the state-space is modeled as an MDP. 
	The second mode narrows the state-space observation to a configurable area around the player. In addition to radius based vision, the POMDP mode also features ray-tracing vision that better mimic the sight of a physical agent. The third and fourth mode is intended for memory research where the agent must find the goal in a limited number of time-steps. In addition to this, the agent is presented with the solution but fades after a few initial time steps. The objective is the for the agent to remember the solution to find the goal. All scenario setups have a variable map-size ranging between $2\times2$ and $56\times56$ tiles.
	
	\subsection{The Deep Line Wars Environment}
	
	The Deep Line Wars environment was first introduced in \cite{Andersen2017}. Deep Line Wars is a real-time strategy environment that makes an extensive state-space reduction to enable swift research in reinforcement learning for RTS games. 
		
	\label{sec:deep_line_wars}
	\begin{figure}
		\centering
		\includegraphics[width=1\linewidth]{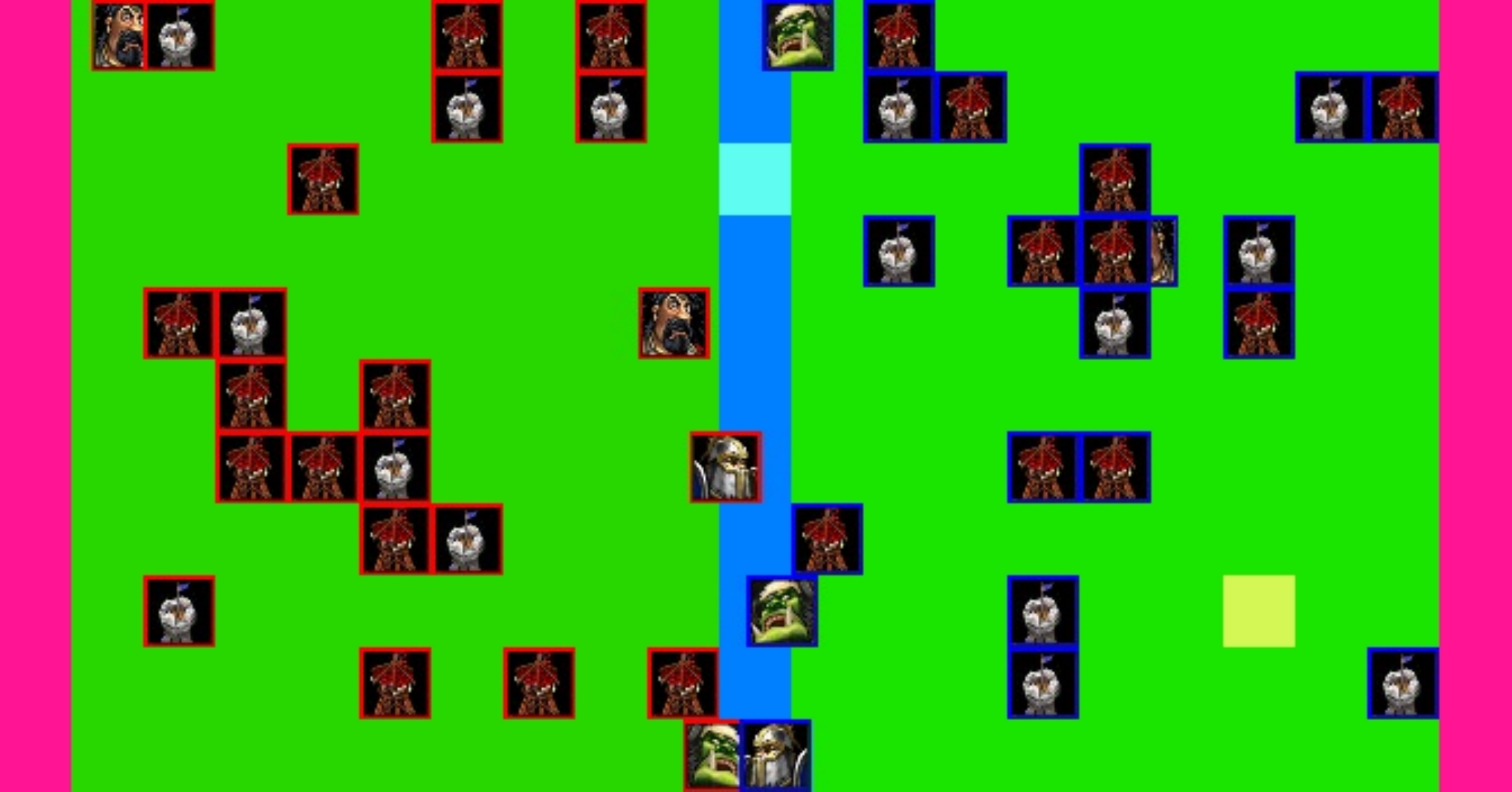}
		\caption{The Graphical User Interface of the Deep Line Wars environment.}
		\label{fig:dlw}
	\end{figure}
	
	The game objective of Deep Line Wars is to invade the enemy player with mercenary units until all health points are depleted, see Figure \ref{fig:dlw}). For every friendly unit that enters the far edge of the enemy base, the enemy health pool is reduced by one. When a player purchases a mercenary unit, it spawns at a random location inside the edge area of the buyers base. Mercenary units automatically move towards the enemy base. To protect the base, players can construct towers that shoot projectiles at the opponents mercenaries. When a mercenary dies, a fair percentage of its gold value is awarded to the opponent. When a player sends a unit, the income is increased by a percentage of the units gold value. As a part of the income system, players gain gold at fixed intervals.

	\section{Experiments}
	\label{sec:experiments}	
	\subsection{Deep Maze Environment Modeling using DVAE}
	The DVAE algorithm must be able to generalize over many similar states to model a vast state-space. DVAE aims to learn the transition function, bringing the state from $s_{t}$ to $s_{t+1} = \mathcal{T}(s_{t}, a_{t})$. We use the deep maze environment because it provides simple rules, with a controllable state-space complexity. Also, we can omit the importance of reward for some scenarios.

	We trained the DVAE model on two \textit{No-Wall Deep Maze} scenarios of size $2\times2$ and $8\times8$. For the encoder and decoder, we used the same convolution architecture as proposed by \cite{Pu2016} and trained for 5000 epochs for $8\times8$ and 1000 epochs for $2\times2$ respectively. For the encoding of actions and states, we concatenated the flattened state-space and action-space, having a fully-connected layer with ReLU activation before calculating the latent-space. We used the Adam optimizer \cite{Kingma2015} with a learning-rate of 1e-08 to update the parameters.

	 \begin{figure}
		\subfloat[A Small, Fully Observable MDP\label{fig:dvae_2x2_loss}]{%
			\includegraphics[width=0.5\linewidth]{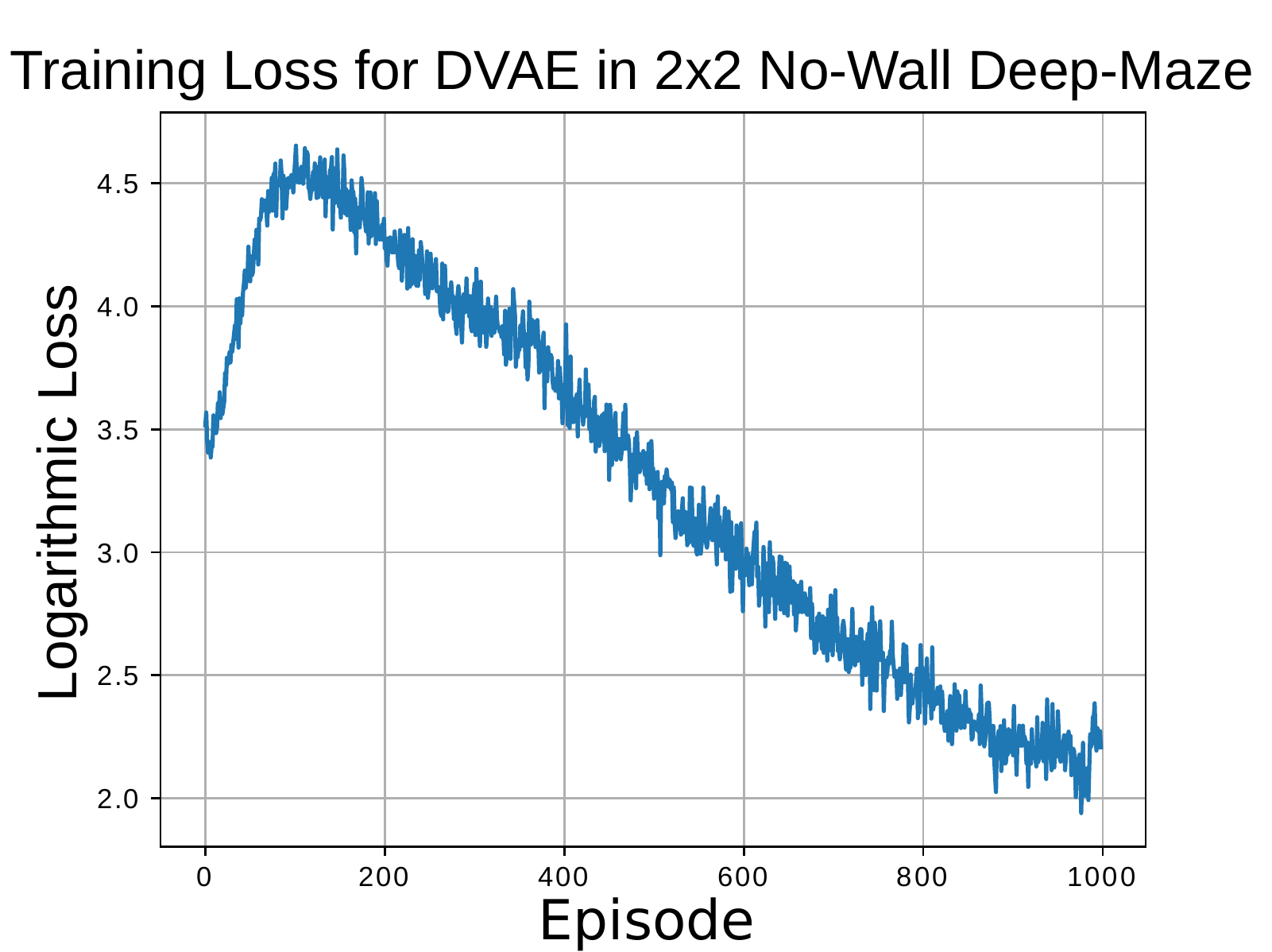}
		}
		\hfill
		\subfloat[A Large, Fully Observable MDP\label{fig:dvae_8x8_loss}]{%
			\includegraphics[width=0.5\linewidth]{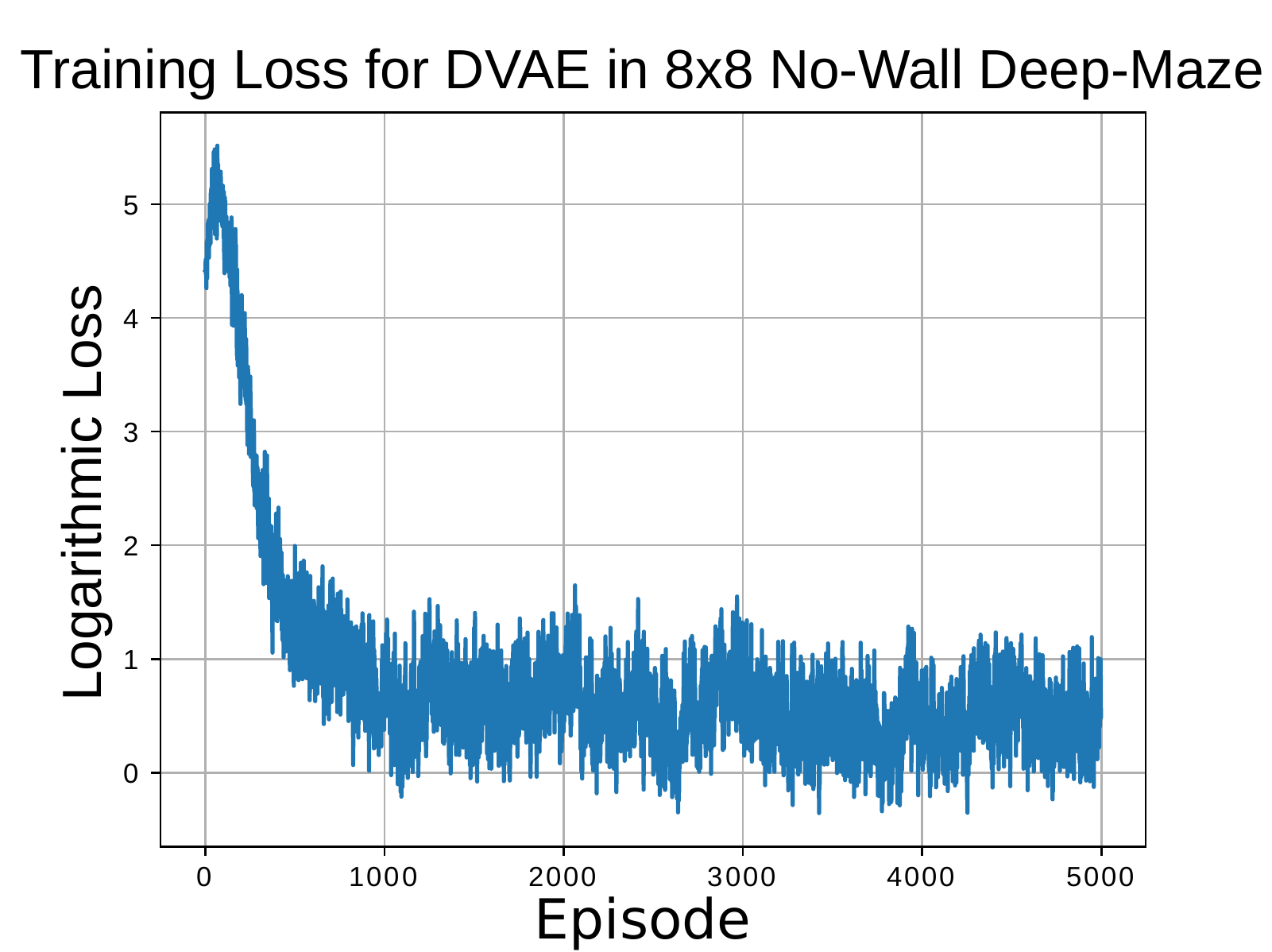}
		}
		\caption{The training loss for DVAE in the $2\times2$ No-Wall and $8\times8$ Deep Maze scenario. The experiment is run for a total of 1000 (5000 for $8\times8$) episodes. The algorithm only trains on 50\% of the state-space to the model for the $2\times2$ environment while the whole state-space is trainable in the $8\times8$ environment.}
		\label{fig:dvae_loss}
	\end{figure}
	
\begin{figure}
		\centering
		\includegraphics[width=1\linewidth]{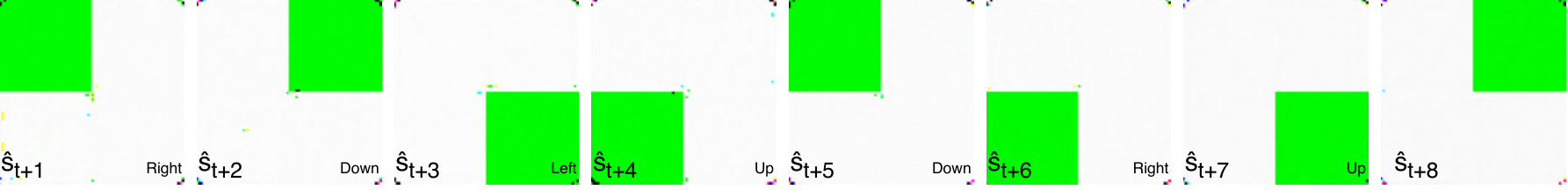}
		\caption{For the $2\times2$ scenario, only 50\% of the environment is explored, leaving artifacts on states where the model is uncertain of the transition function. In more extensive examples, the player disappears, teleports or gets stuck in unexplored areas.}
		\label{fig:deep_maze_2x2}
	\end{figure}

\begin{figure}
		\centering
		\includegraphics[width=1.0\linewidth]{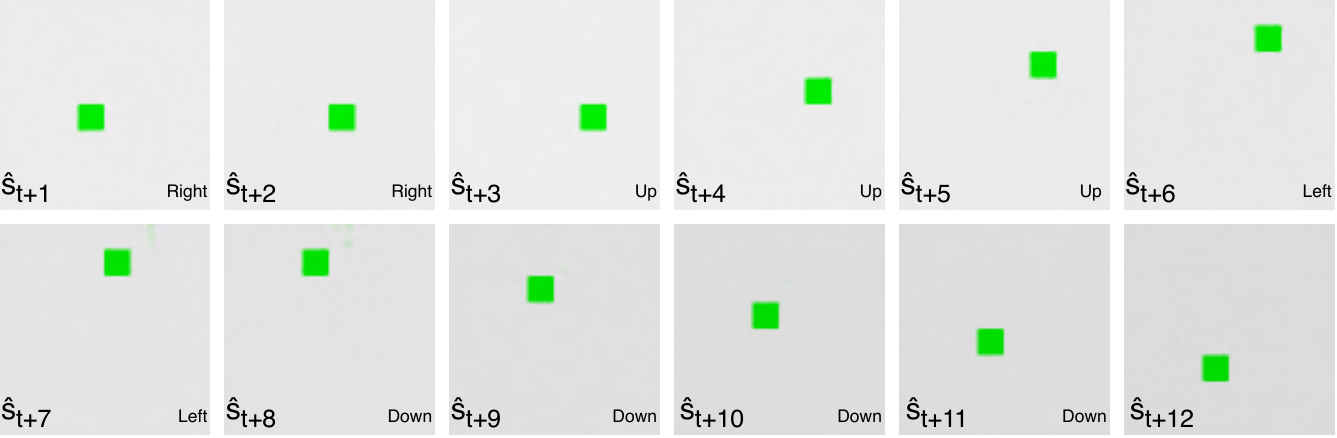}
		\caption{Results of $8\times8$ Deep Maze modeling using the DVAE algorithm. To simplify the environment, no reward signal is received per iteration. The left caption describes current state, $s_{t}$, while the right caption is the action performed to compute, $s_{t+1} = \mathcal{T}(s_{t}, a_{t})$.}
		\label{fig:deep_maze_8x8}
	\end{figure}

	Figure~\ref{fig:dvae_loss} illustrates the loss of the DVAE algorithm in the No-Wall Deep Maze scenario. In the $2\times2$ scenario, DVAE is trained on only 50\% of the state space, which  results in noticeable graphic artifacts in the prediction of future states, see Figure~\ref{fig:deep_maze_2x2}. Because the $8\times8$ environment is fully visible, we see in Figure~\ref{fig:deep_maze_8x8} that the artifacts are exponentially reduced.


			\begin{table}
		\centering
			\caption{Results of the deep maze $11\times11$ and $21\times21$ environment, comparing DQN \cite{Mnih2015}, TRPO \cite{Schulman2015}, and PPO \cite{Schulman2017}. The optimal path yields performance of 100\% while no solution yields 0\%. Each of the algorithms ran 10000 episodes for both map-sizes. The last number represents at which episode the algorithm converged.}
		\begin{tabular}{ | l | l | l |}
			\hline
			Algorithm & Avg Performance $11\times11$& Avg Performance $21\times21$ \\ \hline
			DQN-$\mathcal{\hat{D}}$ & $94.56\%$ @ 9314 & $64.36\%$ @ N/A \\ \hline
			TRPO-$\mathcal{\hat{D}}$ & $96.32\%$  @ 5320& $78.91\%$ @ 7401  \\ \hline
			PPO-$\mathcal{\hat{D}}$ & $98.71\%$ @ 3151 & $89.33\%$ @ 7195 \\ \hline \hline
				
			DQN-$\mathcal{D}$ & $98.26\%$ @ 4314 & $84.63\%$ @ 8241 \\ \hline
			TRPO-$\mathcal{D}$ & $99.32\%$  @ 3320& $92.11\%$ @ 4120  \\ \hline
			PPO-$\mathcal{D}$ & $99.35\%$  @ 2453 & $96.41\%$ @ 2904 \\ \hline
		\end{tabular}

	\label{tbl:perf}
	\end{table}
	


	\subsection{Using $\mathcal{\hat{D}}$ for RL Agents in Deep Maze}
	The goal of this experiment is to observe the performance of RL agents using the generated experience-replay $\mathcal{\hat{D}}$ from Figure~\ref{alg1} in Deep Maze environments of size $11\times11$ and $21\times21$. In Table~\ref{tbl:perf}, we compare the performance of DQN \cite{Mnih2013}, TRPO \cite{Schulman2015}, and PPO \cite{Schulman2017} using the DVAE generated $\mathcal{\hat{D}}$ to tune the parameters.
	
	\begin{figure}[!htp]
		\centering
		\includegraphics[width=1.0\linewidth]{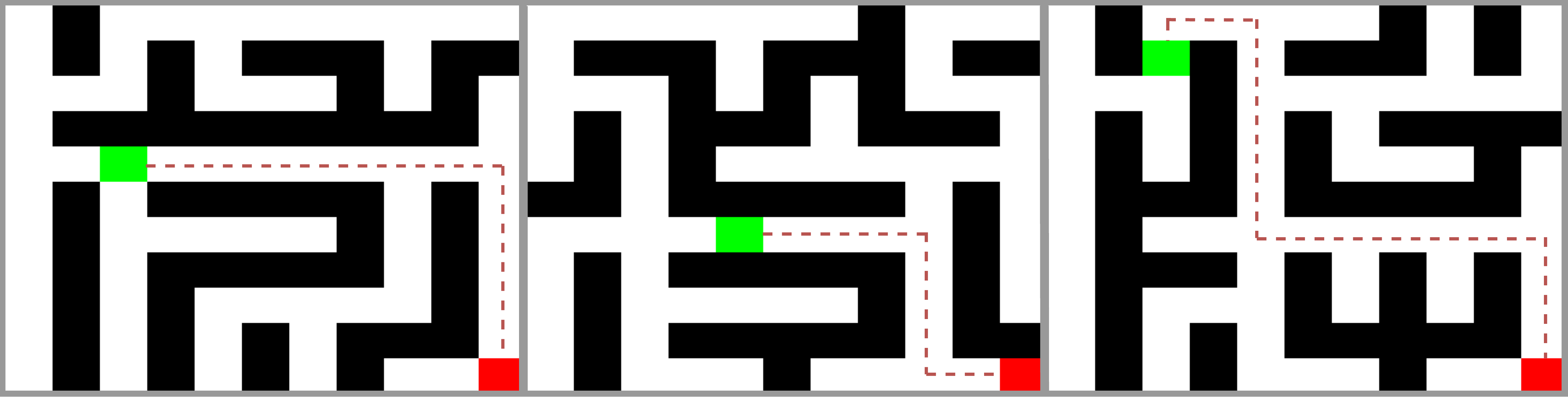}
		\caption{A typical deep maze of size $11\times11$. The lower-right square indicates the goal state, the dotted-line indicates the optimal path, while the final square represents the player's current position in the state-space. The controller agent is DQN, TRPO, and PPO (from left to right).}
		\label{fig:maze_rep_11x11}
	\end{figure}

	Figure \ref{fig:maze_rep_11x11} illustrates three maze variations of size $11\times11$, where the AI has learned the optimal path. We see that the best performing algorithm, PPO  \cite{Schulman2017} beats DQN and TRPO using either $\mathcal{\hat{D}}$ or $\mathcal{D}$. The DQN-$\mathcal{\hat{D}}$ agent did not converge in the $21\times21$ environment, but it is likely that value-based algorithms could struggle with graphical artifacts generated from the DVAE algorithm. These artifacts significantly increase the state-space so that direct-policy algorithms could perform better. 
	
	\subsection{Deep Line Wars Environment Modeling using DVAE}
	The DVAE algorithm works well in more complex environments, such as the Deep Line Wars game environment~\cite{Andersen2017}. Here, we expand the DVAE algorithm with LSTM to improve the interpretation of animations, illustrated  Figure~\ref{fig:dvae_model}. 
	
	\begin{figure}[!htp]
		\centering
		\includegraphics[width=1.0\linewidth]{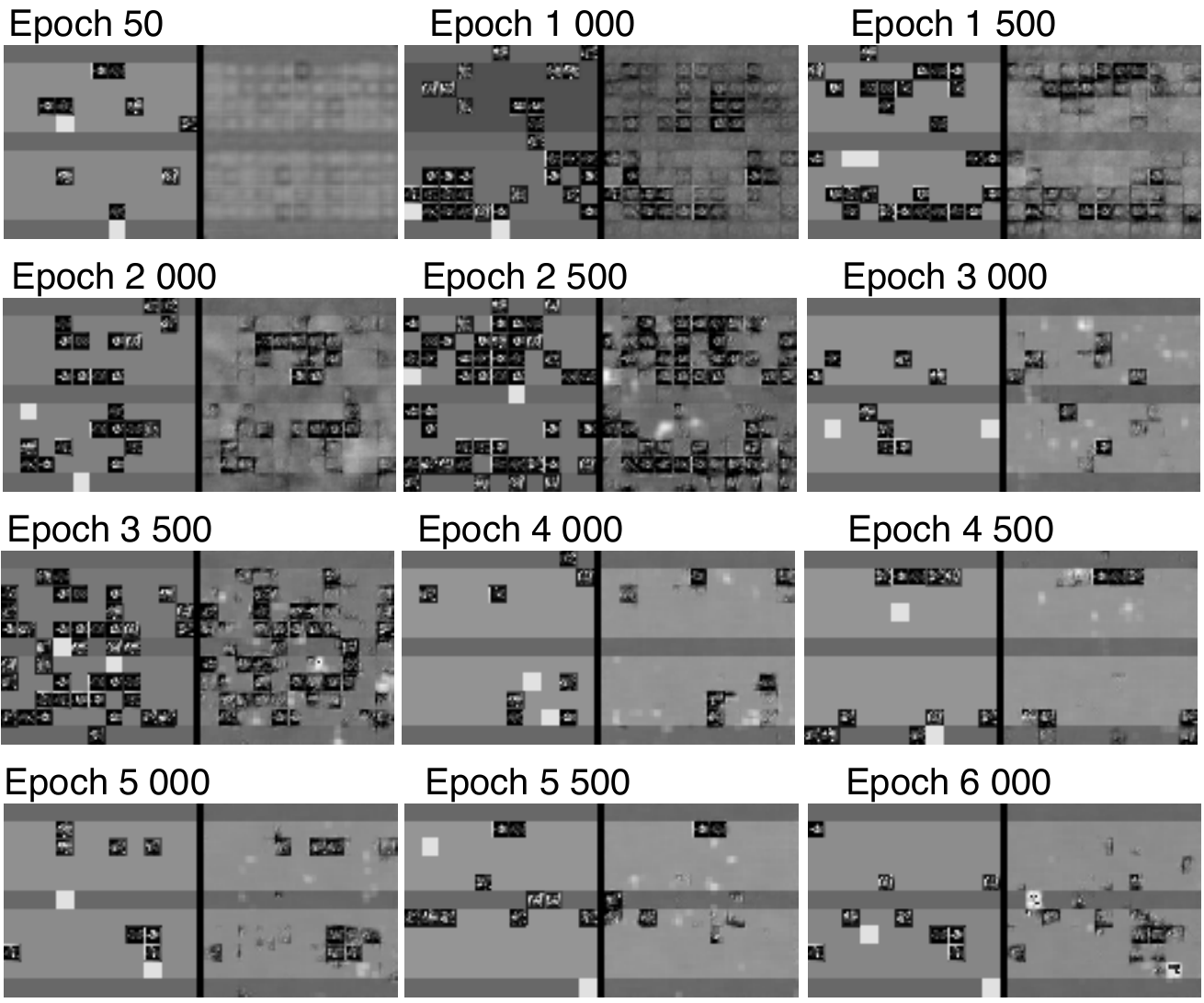}
		\caption{The DVAE algorithm applied to the Deep Line Wars environment. Each epoch illustrates the quality of generated states in the game, where the left image is real state $s$ and the right image is the generated state $\hat{s}$.}
		\label{fig:dlw_dvae}
	\end{figure}
	Figure~\ref{fig:dlw_dvae} illustrates the state quality during training of DVAE in a total of 6000 episodes (epochs). Both players draw actions from a Gaussian distributed policy. The algorithm understands that the player units can be located in any tiles after only 50 epochs, and at 1000 we observe the algorithm makes a more accurate statement of the probability of unit locations (i.e., some units have increased intensity).
	At the end of the training, the DVAE algorithm is to some degree capable of determining both towers, and unit locations at any given time-step during the game episode.

	\section{Conclusion and Future Work}
	\label{sec:conclusion}
	
	
	This paper introduces the \textit{Dreaming Variational Autoencoder} (DVAE) as a neural network based generative modeling architecture to enable exploration in environments with sparse feedback. The DVAE shows promising results in modeling simple non-continuous environments. For continuous environments, such as Deep Line Wars, DVAE performs better using a recurrent neural network architecture (LSTM) while it is sufficient to use only a sequential feed-forward architecture to model non-continuous environments such as Chess, Go, and Deep Maze.
	
	There are, however, several fundamental issues that limit DVAE from fully modeling environments. In some situations, exploration may be a costly act that makes it impossible to explore all parts of the environment in its entirety. DVAE cannot accurately predict the outcome of unexplored areas of the state-space, making the prediction blurry or false.
	
	Reinforcement learning has many unresolved problems, and the hope is that the Deep Maze learning environment can be a useful tool for future research. For future work, we plan to expand the model to model the reward function $\hat{\mathcal{R}}$ using inverse reinforcement learning. DVAE is an ongoing research question, and the goal is that reinforcement learning algorithms could utilize this form of \textit{dreaming} to reduce the need for exploration in real environments.

	\bibliographystyle{splncs04}

\begin{thebibliography}{10}
		\providecommand{\url}[1]{\texttt{#1}}
		\providecommand{\urlprefix}{URL }
		\providecommand{\doi}[1]{https://doi.org/#1}
		
		\bibitem{Andersen2017}
		Andersen, P.A., Goodwin, M., Granmo, O.C.: {Towards a deep reinforcement
			learning approach for tower line wars}. In: Bramer, M., Petridis, M. (eds.)
		Lecture Notes in Computer Science (including subseries Lecture Notes in
		Artificial Intelligence and Lecture Notes in Bioinformatics). vol. 10630
		LNAI, pp. 101--114 (2017)
		
		\bibitem{Arulkumaran2017}
		Arulkumaran, K., Deisenroth, M.P., Brundage, M., Bharath, A.A.: {Deep
			reinforcement learning: A brief survey}. IEEE Signal Processing Magazine
		\textbf{34}(6),  26--38 (2017)
		
		\bibitem{Bangaru2016}
		Bangaru, S.P., Suhas, J., Ravindran, B.: {Exploration for Multi-task
			Reinforcement Learning with Deep Generative Models}. arxiv preprint
		arXiv:1611.09894 (nov 2016)
		
		\bibitem{Blundell2016}
		Blundell, C., Uria, B., Pritzel, A., Li, Y., Ruderman, A., Leibo, J.Z., Rae,
		J., Wierstra, D., Hassabis, D.: {Model-Free Episodic Control}. arxiv preprint
		arXiv:1606.04460  (jun 2016)
		
		\bibitem{Buesing2018}
		Buesing, L., Weber, T., Racaniere, S., Eslami, S.M.A., Rezende, D., Reichert,
		D.P., Viola, F., Besse, F., Gregor, K., Hassabis, D., Wierstra, D.: {Learning
			and Querying Fast Generative Models for Reinforcement Learning}. arxiv
		preprint arXiv:1802.03006  (feb 2018)
		
		\bibitem{Chen2015}
		Chen, K.: {Deep Reinforcement Learning for Flappy Bird}. cs229.stanford.edu
		p.~6 (2015)
		
		\bibitem{Ha2018}
		Ha, D., Schmidhuber, J.: {World Models}. arxiv preprint arXiv:1803.10122  (mar
		2018)
		
		\bibitem{Higgins2016}
		Higgins, I., Matthey, L., Pal, A., Burgess, C., Glorot, X., Botvinick, M.,
		Mohamed, S., Lerchner, A.: {beta-VAE: Learning Basic Visual Concepts with a
			Constrained Variational Framework}. International Conference on Learning
		Representations  (nov 2016)
		
		\bibitem{Higgins2017}
		Higgins, I., Pal, A., Rusu, A., Matthey, L., Burgess, C., Pritzel, A.,
		Botvinick, M., Blundell, C., Lerchner, A.: {DARLA: Improving Zero-Shot
			Transfer in Reinforcement Learning}. In: Precup, D., Teh, Y.W. (eds.)
		Proceedings of the 34th International Conference on Machine Learning.
		Proceedings of Machine Learning Research, vol.~70, pp. 1480--1490. PMLR,
		International Convention Centre, Sydney, Australia (2017)
		
		\bibitem{Kaelbling1996}
		Kaelbling, L.P., Littman, M.L., Moore, A.W.: {Reinforcement Learning: A
			Survey}. Journal of Artificial Intelligence Research  (apr 1996)
		
		\bibitem{Kingma2015}
		Kingma, D.P., Ba, J.L.: {Adam: A Method for Stochastic Optimization}.
		Proceedings, International Conference on Learning Representations 2015
		(2015)
		
		\bibitem{Li2017}
		Li, Y.: {Deep Reinforcement Learning: An Overview}. arxiv preprint
		arXiv:1701.07274  (jan 2017)
		
		\bibitem{Mnih2016}
		Mnih, V., Badia, A.P., Mirza, M., Graves, A., Lillicrap, T., Harley, T.,
		Silver, D., Kavukcuoglu, K.: {Asynchronous Methods for Deep Reinforcement
			Learning}. In: Balcan, M.F., Weinberger, K.Q. (eds.) Proceedings of The 33rd
		International Conference on Machine Learning. Proceedings of Machine Learning
		Research, vol.~48, pp. 1928--1937. PMLR, New York, New York, USA (2016)
		
		\bibitem{Mnih2013}
		Mnih, V., Kavukcuoglu, K., Silver, D., Graves, A., Antonoglou, I., Wierstra,
		D., Riedmiller, M.: {Playing Atari with Deep Reinforcement Learning}. Neural
		Information Processing Systems  (dec 2013)
		
		\bibitem{Mnih2015}
		Mnih, V., Kavukcuoglu, K., Silver, D., Rusu, A.A., Veness, J., Bellemare, M.G.,
		Graves, A., Riedmiller, M., Fidjeland, A.K., Ostrovski, G., Petersen, S.,
		Beattie, C., Sadik, A., Antonoglou, I., King, H., Kumaran, D., Wierstra, D.,
		Legg, S., Hassabis, D.: {Human-level control through deep reinforcement
			learning}. Nature  \textbf{518}(7540),  529--533 (feb 2015)
		
		\bibitem{Mousavi2018}
		Mousavi, S.S., Schukat, M., Howley, E.: {Deep Reinforcement Learning: An
			Overview}. In: Bi, Y., Kapoor, S., Bhatia, R. (eds.) Proceedings of SAI
		Intelligent Systems Conference (IntelliSys) 2016. pp. 426--440. Springer
		International Publishing, Cham (2018)
		
		\bibitem{Pu2016}
		Pu, Y., Gan, Z., Henao, R., Yuan, X., Li, C., Stevens, A., Carin, L.:
		{Variational Autoencoder for Deep Learning of Images, Labels and Captions}.
		In: Lee, D.D., Sugiyama, M., Luxburg, U.V., Guyon, I., R., G. (eds.) Advances
		in Neural Information Processing Systems. pp. 2352--2360. Curran Associates,
		Inc. (2016)
		
		\bibitem{Schulman2015}
		Schulman, J., Levine, S., Abbeel, P., Jordan, M., Moritz, P.: {Trust Region
			Policy Optimization}. In: Bach, F., Blei, D. (eds.) Proceedings of the 32nd
		International Conference on Machine Learning. Proceedings of Machine Learning
		Research, vol.~37, pp. 1889--1897. PMLR, Lille, France (2015)
		
		\bibitem{Schulman2017}
		Schulman, J., Wolski, F., Dhariwal, P., Radford, A., Klimov, O.: {Proximal
			Policy Optimization Algorithms}. arxiv preprint arXiv:1707.06347  (jul 2017)
		
		\bibitem{Sutton1998}
		Sutton, R.S., Barto, A.G.: {Reinforcement Learning: An Introduction}, vol.~9.
		MIT Press (1998)
		
		\bibitem{VanSeijen2017}
		{Van Seijen}, H., Fatemi, M., Romoff, J., Laroche, R., Barnes, T., Tsang, J.:
		{Hybrid Reward Architecture for Reinforcement Learning}. In: Guyon, I.,
		Luxburg, U.V., Bengio, S., Wallach, H., Fergus, R., Vishwanathan, S.,
		Garnett, R. (eds.) Advances in Neural Information Processing Systems 30, pp.
		5392--5402. Curran Associates, Inc. (2017)
		
		\bibitem{Xiao2016}
		Xiao, T., Kesineni, G.: {Generative Adversarial Networks for Model Based
			Reinforcement Learning with Tree Search}. Tech. rep., University of
		California, Berkeley (2016)
		
	\end{thebibliography}

\end{document}